\documentclass[11pt,a4paper]{article}
\usepackage[dvipsnames]{xcolor}
\usepackage[hyperref]{acl2018}

\usepackage[utf8]{inputenc}
\usepackage[T1]{fontenc}

\usepackage{times}
\usepackage{latexsym}
\usepackage{booktabs, tabularx}
\usepackage{amsmath}
\usepackage{url}
\usepackage{todonotes}
\usepackage{soul}

\usepackage[linesnumbered,lined,commentsnumbered]{algorithm2e}

\aclfinalcopy % Uncomment this line for the final submission
\title{A Large-Scale Test Set for the Evaluation of Context-Aware Pronoun Translation in Neural Machine Translation}

\makeatletter
\def\blfootnote{\xdef\@thefnmark{}\@footnotetext}
\makeatother

\author{Mathias M\"{u}ller$^{1,2}$ \quad Annette Rios$^1$ \quad Elena Voita$^{3,4}$ \quad Rico Sennrich$^{1,5}$ \bigskip\\
  $^1$Institute of Computational Linguistics, University of Zurich \smallskip\\
  $^2$Amazon Research, Berlin$^*$ \smallskip\\
  $^3$Yandex Research, Russia\quad
  $^4$University of Amsterdam, Netherlands \smallskip\\
  $^5$School of Informatics, University of Edinburgh
	}
\date{} 

\begin{document}
\maketitle
\blfootnote{$^*$ Work performed prior to joining Amazon.}

\begin{abstract}
  The translation of pronouns presents a special challenge to machine translation to this day, since it often requires context outside the current sentence. Recent work on models that have access to information across sentence boundaries has seen only moderate improvements in terms of automatic evaluation metrics such as BLEU. 
However, metrics that quantify the overall translation quality are ill-equipped to measure gains from additional context. We argue that a different kind of evaluation is needed to assess how well models translate inter-sentential phenomena such as pronouns. This paper therefore presents a test suite of contrastive translations focused specifically on the translation of pronouns.
Furthermore, we perform experiments with several context-aware models. We show that, while gains in BLEU are moderate for those systems, they outperform baselines by a large margin in terms of accuracy on our contrastive test set. Our experiments also show the effectiveness of parameter tying for multi-encoder architectures.

\end{abstract}

\section{Introduction}

Even though machine translation has improved considerably with the advent of neural machine translation (NMT) \citep{Sutskever2014,Bahdanau2015}, the translation of pronouns remains a major issue. They are notoriously hard to translate since they often require context outside the current sentence.

As an example, consider the sentences in Figure \ref{fig:ecb}. In both languages, there is a pronoun in the second sentence that refers to the European Central Bank. When the second sentence is translated from English to German, the translation of the pronoun \textit{it} is ambiguous. This ambiguity can only be resolved with context awareness: if a translation system has access to the previous English sentence, the previous German translation, or both, it can determine the antecedent the pronoun refers to. In this German sentence, the antecedent \textit{Europ\"{a}ische Zentralbank} dictates the feminine gender of the pronoun \textit{sie}.

\begin{figure}
\begin{itemize}
\item[\textbf{EN}] However, \ul{the European Central Bank (ECB)} took an interest in it. \textit{It} describes bitcoin as ``the most successful virtual currency''.
\end{itemize}

\begin{itemize}
\item[\textbf{DE}] Dennoch hat \ul{die Europ\"{a}ische Zentralbank (EZB)} Interesse hierf\"{u}r gezeigt. \textit{Sie} beschreibt Bitcoin als ``die virtuelle W\"{a}hrung mit dem gr\"{o}ssten Erfolg''.
\end{itemize} 

\caption{Example sentence illustrating how the translation of pronouns is ambiguous on a sentence level. Pronouns of interest are in italics, and the antecedents they refer to are underlined. Taken from WMT \texttt{newstest2013}.} \label{fig:ecb}
\end{figure}

It is unfortunate, then, that current NMT systems generally operate on the sentence level \citep{Vaswani2017,Gehring2017,Hieber2017}. Documents are translated sentence-by-sentence for practical reasons, such as line-based processing in a pipeline and reduced computational complexity. Furthermore, improvements of larger-context models over baselines in terms of document-level metrics such as BLEU or RIBES have been moderate, so that their computational overhead does not seem justified, and so that it is hard to develop more effective context-aware architectures and empirically validate them.

To address this issue, we present an alternative way of evaluating larger-context models on a test set that allows to specifically  measure a model's capability to correctly translate pronouns. 
The test suite consists of pairs of source and target sentences, in combination with contrastive translation variants (for evaluation by model scoring) and additional linguistic and contextual information (for further analysis). The resource is freely available.\footnote{\url{https://github.com/ZurichNLP/ContraPro}}
Additionally, we evaluate several context-aware models that have recently been proposed in the literature on this test set, and extend existing models with parameter tying.

The main contributions of our paper are:

\begin{itemize}
\item We present a large-scale test set to evaluate the accuracy with which NMT models translate the English pronoun {\em it} to its German counterparts {\em es}, {\em sie} and {\em er}.
\item We evaluate several context-aware systems and show how targeted, contrastive evaluation is an effective tool to measure improvement in pronoun translation.
\item We empirically demonstrate the effectiveness of parameter tying in multi-encoder context-aware models.
\end{itemize}

Section \ref{sec:related-work} explains how our paper relates to existing work on context-aware models and the evaluation of pronoun translation. Section \ref{sec:testset} describes our test suite. The context-aware models we use in our experiments are detailed in Section \ref{sec:models}. We discuss our experiments in Section \ref{sec:experiments} and the results in Section \ref{sec:evaluation}.

\section{Related Work} \label{sec:related-work}

Two lines of work are related to our paper: research on context-aware translation (described in Section \ref{subsec:related-models}) and research on focused evaluation of pronoun translation (described in Section \ref{subsec:related-eval}).

\subsection{Context-Aware NMT Models} \label{subsec:related-models}

If the translation of a pronoun requires context beyond the current sentence (see the example in Figure \ref{fig:ecb}), a natural extension of sentence-level NMT models is to condition the model prediction on this necessary context. In the following, we describe a number of existing approaches to making models ``aware'' of additional context.

\begin{table*}
\begin{center}
\begin{tabular}{lcccccc}
& \multicolumn{2}{c}{\textbf{Languages}} & \multicolumn{4}{c}{\textbf{Context types}} \\
& source & target & source & target & preceding & following \\
\hline
\citet{Tiedemann2017} & DE & EN & x & x & x & \\
\citet{Jean2017} & EN & FR/DE & x &  & x & \\
\citet{Wang2017} & ZH & EN & x &  & x & \\
\citet{Voita2018} & EN & RU & x &  & x & x\\
\citet{Bawden2017} & EN & FR & x & x & x & \\
\citet{DBLP:conf/acl/HaffariM18} & FR/DE/ET & EN & x & x & x & \\
\citet{agrawal2018} & EN & IT & x & x & x & x \\
\end{tabular}
\caption{Overview of context-aware translation models in related work.}\label{tbl:rel-overview}
\end{center}
\end{table*}

The simplest possible extension is to translate units larger than sentences. \citet{Tiedemann2017} concatenate each sentence with the sentence that precedes it, for the source side of the corpus or both sides.
All of their models are standard sequence-to-sequence models built with recurrent neural networks (RNNs), since the method does not require any architectural change. \citet{agrawal2018} use the same concatenation technique with a Transformer architecture \cite{Vaswani2017}, and experiment with wider context.

A number of works do propose changes to the NMT architecture. A common technique is to extend a standard encoder-decoder model by additional encoders for the context sentence(s), with a modified attention mechanism \cite{Jean2017,Bawden2017,Voita2018}. One aspect that differs between these works is the architecture of the encoder and attention. While \citet{Jean2017,Bawden2017} extend an RNN encoder-decoder with a second encoder that the decoder attends to, \citet{Voita2018} extend the Transformer architecture with an encoder that is attended to by the main encoder. \citet{Voita2018} also introduce parameter sharing between the main encoder and the context encoder, but do not empirically demonstrate its importance.

While the number of encoded sentences in the previous work is fixed, \citet{Wang2017,DBLP:conf/acl/HaffariM18} explore the integration of variable-size context through a hierarchical architecture, where a first-level RNN reads in words to produce sentence vectors, which are then fed into a second-level RNN to produce a document summary.

Apart from differences in the architectures, related work varies in whether it considers source context, target context, or both (see Table \ref{tbl:rel-overview} for an overview of language arcs and context types).
Some work considers only source context, but for pronoun translation, target-side context is intuitively important for disambiguation, especially if the antecedent itself is ambiguous.
In our evaluation, we therefore emphasize models that take into account both source and target context.

Our experiments are based on models from \citet{Bawden2017}, who have released their source code.\footnote{\url{https://github.com/rbawden/nematus}} We extend their models with parameter sharing, which was shown to be beneficial by \citet{Voita2018}. Additionally, we consider a concatenative baseline, similar to \citet{Tiedemann2017}, and Transformer-based models \citep{Voita2018}.

\subsection{Evaluation of Pronoun Translation} \label{subsec:related-eval}

Pronouns can serve a variety of functions with complex cross-lingual variation \cite{guillouphd},
and hand-picked, manually annotated test suites have been presented for the evaluation of pronoun translation \cite{Guillou2016b,Isabelle2017,Bawden2017}.
While suitable for analysis, the small size of the test suites makes it hard to make statistically confident comparisons between systems,
and the hand-picked nature of the test suites introduces biases.\footnote{For example, all pronoun examples in the test suite by \citet{Bawden2017} require the previous target sentence for disambiguation, and thus do not reward models that condition on more than one sentence of context.}
To overcome these problems, we opted for a fully automatic approach to constructing a large-scale test suite.

Conceptually, our test set is most similar to the ``cross-lingual pronoun prediction'' task held at DiscoMT and WMT in recent years \cite{Hardmeier2015,Guillou2016,Loaiciga2017}: participants are asked to fill a gap in a target sentence, where gaps correspond to pronouns.

The first edition of the task focused on English$\to$French, and it was found that local context (such as the verb group) was a strong signal for pronoun prediction. Hence, future editions only provided target-side lemmas instead of fully inflected forms, which makes the task less suitable to evaluate
end-to-end neural machine translation systems, although such systems have been trained on the task \cite{jean-EtAl:2017:DiscoMT}.

\citet{Loaiciga2017} do not report on the proportion of intra-sentential and inter-sentential anaphora in their test set, but the two top-performing systems only made use of intra-sentential information.
Our test suite focuses on allowing the comparison of end-to-end context-aware NMT systems, and we thus extract a large number of \emph{inter-sentential anaphora}, with meta-data allowing for a focus on inter-sentential anaphora with a long distance between the pronoun and its antecedent.
Our focus on evaluating end-to-end NMT systems also relieves us from having to provide annotated training sets, and reduces pressure to achieve balance and full coverage of phenomena.\footnote{For example, we do not consider cases where English {\em it} is translated into something other than a personal pronoun. While this would be a severe blind spot in a training set for pronoun prediction, the focused nature of our test suite does not impair the performance of end-to-end NMT systems on other phenomena.}

An alternative approach to automatically evaluate pronoun translation are reference-based methods that produce a score based on word alignment between source, translation output, and reference translation, and identification of pronouns in them, such as AutoPRF \cite{hardmeier2010} and APT \cite{miculicich2017validation}.
\citet{guillou2018} perform a human meta-evaluation and show substantial disagreement between reference-based metrics and human judges, especially because there often exist valid alternative translations that use different pronouns than the reference.
Our test set, and our protocol of generating contrastive examples, is focused on selected pronouns to minimize the risk of producing contrastive examples that are actually valid translations.

\section{Test set with contrastive examples} \label{sec:testset}

Contrastive evaluation requires a large set of suitable examples that involve the translation of pronouns. As additional goals, our test set is designed to 1) focus on \textit{hard} cases, so that it can be used as a benchmark to track progress in context-aware translation and 2) allow for fine-grained analysis.

Section \ref{subsec:automatic-extraction} describes how we extract our data set. Section \ref{subsec:eval-scoring} explains how, given a set of contrastive examples, contrastive evaluation works.

\subsection{Automatic extraction of contrastive examples from corpora} \label{subsec:automatic-extraction}

We automatically create a test set from the OpenSubtitles corpus \citep{Lison2016}.\footnote{\url{http://opus.nlpl.eu/OpenSubtitles2016.php}} The goal is to provide a large number of difficult test cases where an English pronoun has to be translated to a German pronoun.

The most challenging cases are translating {\em it} to either {\em er, sie} or {\em es}, depending on the grammatical gender of the antecedent.\footnote{The pronouns {\em he} and {\em she} usually refer to a person in English, and since persons do not change gender in the translation, we assume that learning the correspondences {\em he} $\rightarrow$ {\em er} and {\em she} $\rightarrow$ {\em sie} does not present a challenge for a model. Cases where {\em he} or {\em she} refer to a noun that is not a person are possible, but extremely rare.} Not only is the translation of {\em it} ambiguous, there is also class imbalance in the training data (see Table \ref{Tab:fast_align_freqs}). There is roughly a 30\% probability that {\em it} is aligned to {\em es},\footnote{Note that these statistics include non-referential uses of {\em it}, that we exclude from our testset.} which makes it difficult to learn to translate {\em er} and {\em sie}. We use parsing and automatic co-reference resolution to find translation pairs that satisfy our constraints.

\begin{table}
\begin{center}
\begin{tabular}{lrr}
\toprule
Alignment & Frequency &  Probability \\
\midrule
it$\rightarrow$es & 255764 & 0.334 \\
it$\rightarrow$sie & 64446 &  0.084 \\
it$\rightarrow$er & 44543  & 0.058 \\
it$\rightarrow$ist & 42614 &  0.055 \\
it$\rightarrow$Sie & 26054 &  0.034 \\
it$\rightarrow$,  & 21037 &  0.027 \\
it$\rightarrow$das & 17992 &  0.023 \\
it$\rightarrow$dies  &  11943 &  0.015 \\
it$\rightarrow$wird  &  11886 &  0.015 \\
it$\rightarrow$man & 10539 &  0.013 \\
it$\rightarrow$ihn & 7744  &  0.010 \\
\bottomrule
\end{tabular}
\caption{Frequency and probability of alignments of \textit{it} in the training data of our systems (all data from the WMT 2017 news translation task). Alignments are produced by a fast\_align model.}\label{Tab:fast_align_freqs}
\end{center}
\end{table}

To provide a basis for filtering with constraints, we tokenize the whole data set with the Moses tokenizer, generate symmetric word alignments with fast\_align \citep{Dyer2013}, parse the English text with CoreNLP \citep{Manning2014}, parse the German text with ParZu \citep{sennrich2013} and perform coreference resolution on both sides. The coreference chains are obtained with the neural model of CoreNLP for English, and with CorZu for German \citep{Tuggener2016}, respectively.

\begin{table*}
\begin{center}
\begin{tabular}{lll}
\toprule
source: & {\em It could get tangled in your hair.}\\
reference: & {\em \textbf{Sie} k\"onnte sich in deinem Haar verfangen.} \\
\midrule
contrastive:& {\em \textbf{Er} k\"onnte sich in deinem Haar verfangen.} \\
contrastive:& {\em \textbf{Es} k\"onnte sich in deinem Haar verfangen.} \\
\midrule
antecedent en: & a bat \\
antecedent de: & eine Fledermaus (f.) \\
antecedent distance : & 1 \\
\bottomrule
\end{tabular}
\caption{Example sentence pair with contrastive translations. An antecedent distance of 1 means that the antecedent is in the immediately preceding sentence.}\label{Tab:example}
\end{center}
\end{table*}

Then we opt for high-precision, aggressive filtering, according to the following protocol: for each pair of sentences $(e, f)$ in English and German, extract iff

\begin{itemize}
\item $e$ contains the English pronoun {\em it}, and $f$ contains a German pronoun that is third person singular ({\em er, sie } or {\em es}), as indicated by their part-of-speech tags;
\item those pronouns are aligned to each other;
\item both pronouns are in a coreference chain;
\item their nominal antecedents in the coreference chain are aligned on word level.
\end{itemize}

This removes most candidate pairs, but is necessary to overcome the noise introduced by our preprocessing pipeline, most notably coreference resolution.
From the filtered set, we create a balanced test set by randomly sampling 4000 instances of each of the three translations of {\em it} under consideration ({\em er}, {\em sie}, {\em es}). We do not balance antecedent distance. See Table \ref{Tab:testsetStats} for the distribution of pronoun pairs and antecedent distance in the test set.

For each sentence pair in the resulting test set, we introduce \textit{contrastive translations}.
A contrastive translation is a translation variant where the correct pronoun is swapped with an incorrect one. For an example, see Table \ref{Tab:example}, where the pronoun {\em it} in the original translation corresponds to {\em sie} because the antecedent {\em bat} is a feminine noun in German ({\em Fledermaus}). We produce wrong translations by replacing {\em sie} with one of the other pronouns ({\em er}, {\em es}).

Note that, by themselves, these contrastive translations are grammatically correct if the antecedent is outside the current sentence. The test set also contains pronouns with an antecedent in the same sentence (antecedent distance 0). Those examples do not require any additional context for disambiguation and we therefore expect the sentence-level baseline to perform well on them.

We take extra care to ensure that the resulting contrastive translations are grammatically correct, because ungrammatical sentences are easily dismissed by an NMT system. For instance, if there are any possessive pronouns (such as {\em seine}) in the sentence, we also change their gender to match the personal pronoun replacement.

The German coreference resolution system does not resolve {\em es} because most instances of {\em es} in German are either non-referential forms, or they refer to a clause instead of a nominal antecedent. We limit the test set to nominal antecedents, as these are the only ambiguous cases with respect to translation.
For this reason, we have to rely entirely on the English coreference links for the extraction of sentence pairs with {\em it$\rightarrow$es}, as opposed to pairs with {\em it$\rightarrow$er} and {\em it$\rightarrow$sie} where we have coreference chains in both languages.\footnote{There are some cases where the antecedent is listed as {\em it} in the test set. This is our fallback behaviour if the coreference chain does not contain any noun. In that case, we do not know the true antecedent.}

Our extraction process respects document boundaries, to ensure we always search for the right context. We extract additional information from the annotated documents, such as the distance (in sentences) between pronouns and their antecedents, the document of origin, lemma, morphology and dependency information if available.

\begin{table}
\begin{center}
\begin{tabular}{crrrr}
\toprule
distance & {\em it}$\rightarrow${\em es} & {\em it}$\rightarrow${\em er} & {\em it}$\rightarrow${\em sie} &total \\ \cmidrule{2-5}
 
\phantom{$>$}0 & 872 & 736 & 792 & 2400\\
\phantom{$>$}1 & 1892 & 2577 & 2606& 7075\\
\phantom{$>$}2 & 631 & 459 & 420 &1510\\
\phantom{$>$}3 & 274 & 167 & 132 & 573\\
$>$3 & 331 & 61 & 50 & 442\\ \cmidrule{2-5}
total & 4000 & 4000 & 4000 & 12000\\
\bottomrule

\end{tabular}
\caption{Test set frequencies of pronoun pairs and antecedent distance (measured in sentences).}\label{Tab:testsetStats}
\end{center}
\end{table}

\subsection{Evaluation by scoring} \label{subsec:eval-scoring}

Contrastive evaluation is different from conventional evaluation of machine translation in that it does not require any translation. Rather than testing a model's ability to translate, it is a method to test a model's ability to \textit{discriminate} between given good and bad translations.

We exploit the fact that NMT systems are in fact language models of the target language, conditioned on source text. Like language models, NMT systems can be used to compute a model score (the negative log probability) for an existing translation. Contrastive evaluation, then, means to compare the model score of two pairs of inputs: $(actual\ source,\ reference\ translation)$ and $(actual\ source,\ contrastive\ translation)$. If the model score of the actual reference translation is higher, we assume that this model can detect wrong pronoun translations.

However, this does \textit{not} mean that systems actually produce the reference translation when given the source sentence for translation. An entirely different target sequence might rank higher in the system's beam during decoding. The only conclusion permitted by contrastive evaluation is whether or not the reference translation is more probable than a contrastive variant.

If the model score of the reference is indeed higher, we refer to this outcome as a ``correct decision'' by the model. The model's decision is only correct if the reference translation has a higher score than any contrastive translation. In our evaluation, we aggregate model decisions on the whole test set and report the overall percentage of correct decisions as accuracy.

During scoring, the model is provided with reference translations as target context, while during translation, the model needs to predict the full sequence. It is an open question to what extent performance deteriorates when context is itself predicted, and thus noisy. We highlight that the same problem arises for sentence-level NMT, and has been addressed with alternative training strategies \cite{Ranzato2015}.

\section{Context-Aware NMT Models} \label{sec:models}
This section describes several context-aware NMT models that we use in our experiments. They fall into two major categories: models based on RNNs and models based on the Transformer architecture \cite{Vaswani2017}. We experiment with additional context on the source side and target side.

\subsection{Recurrent Models}

We consider the following recurrent baselines:

\textbf{baseline} Our baseline model is a standard bidirectional RNN model with attention, trained with Nematus. It operates on the sentence level and does not see any additional context. The input and output embeddings of the decoder are tied, encoder embeddings are not.

\textbf{concat22} We concatenate each sentence with one preceding sentence, for both the source and target side of the corpus. Then we train on this new data set without any changes to the model architecture. This very simple method is inspired by \citet{Tiedemann2017}.

The following models are taken, or slightly adapted, from \citet{Bawden2017}. For this reason, we give only a very short description of them here and the reader is referred to their work for details.

\textbf{s-hier} A multi-encoder architecture with hierarchical attention. This model has access to one additional context: the previous source sentence. It is read by a separate encoder, and attended to by an additional attention network. The output of the resulting two attention vectors is combined with yet another attention network.

\textbf{s-t-hier} Identical to {\em s-hier}, except that it considers two additional contexts: the previous source sentence and previous target sentence. Both are read by separate encoders, and sequences from all encoders are combined with hierarchical attention.

\textbf{s-hier-to-2} The model has an additional encoder for source context, whereas the target side of the corpus is concatenated, in the same way as for {\em concat22}. This model achieved the best results in \citet{Bawden2017}.

For each variant, we also introduce and test weight tying: we share the parameters of embedding matrices between encoders that read the same kind of text (source or target side). 

\subsection{Transformer Models}

All remaining models are based on the Transformer architecture \cite{Vaswani2017}. A Transformer avoids recurrence completely: it follows an encoder-decoder architecture using stacked self-attention and fully connected layers for both the encoder and decoder.

\textbf{baseline} A standard context-agnostic Transformer. All model parameters are identical to a {\em Transformer-base} in \citet{Vaswani2017}.

\textbf{concat22} A simple concatentation model where only the training data is modified, in the same way as for the recurrent {\em concat22} model.

\textbf{concat21} Trained on data where the preceding sentence is concatenated to the current one only on the source side. This model is also taken from \citet{Tiedemann2017}.

\textbf{\citet{Voita2018}} A more sophisticated context-aware Transformer that uses source context only. It has a separate encoder for source context, but all layers except the last one are shared between encoders. A source and context sentence are first encoded independently, and then a single attention layer and a gating function are used to produce a context-aware representation of the source sentence. Such restricted interaction with context is shown to be beneficial for analysis of contextual phenomena captured by the model. For details the reader is referred to their work.

\section{Experiments} \label{sec:experiments}

We train all models on the data from the WMT 2017 English$\to$German news translation shared task ($\sim$ 5.8 million sentence pairs). These corpora do not have document boundaries, therefore a small fraction of sentences will be paired with wrong context,
but we expect the model to be robust against occasional random context (see also \citealt{Voita2018}). Experimental setups for the RNN and Transformer models are different, and we describe them separately.

All RNN-based models are trained with Nematus \cite{sennrich-EtAl:2017:EACLDemo}. We learn a joint BPE model with 89.5k merge operations \cite{sennrich-haddow-birch:2016:P16-12}. We train shallow models with an embedding size of 512, a hidden layer size of 1024 and layer normalization. Models are trained with Adam \cite{Kingma2015}, with an initial learning rate of 0.0001. We apply early stopping based on validation perplexity. The batch size for training is 80, and the maximum length of training sequences is 100 (if input sentences are concatenated) or 50 (if input lines are single sentences).

For our Transformer-based experiments, we use a custom implementation and follow the hyperparameters from \citet{Vaswani2017,Voita2018}.
Systems are trained on lowercased text that was encoded using BPE (32k merge operations).
Models consist of 6 encoder and decoder layers with 8 attention heads.
The hidden state size is 512, the size of feedforward layers is 2048.

Model performance is evaluated in terms of BLEU, on \texttt{newstest2017}, \texttt{newstest2018} and all sentence pairs from our pronoun test set. We compute scores with SacreBLEU \cite{post2018call}.\footnote{Our (cased) SacreBLEU signature is \texttt{BLEU+c.mixed+ l.en-de+\#.1+s.exp+t.wmt\{17,18\}+tok.13a+ v.1.2.10}.} Evaluation with BLEU is done mainly to control for overall translation quality.

To evaluate pronoun translation, we perform contrastive evaluation and report the accuracy of models on our contrastive test set.

\begin{table*}
\centering
\begin{tabular}{lcccccc}
\toprule
& \multicolumn{2}{c}{newstest2017} & \multicolumn{2}{c}{newstest2018} & \multicolumn{2}{c}{pronoun set} \\
& cased & uncased & cased & uncased & cased & uncased \\
\cmidrule{2-7}
baseline & 23.0 & 23.7  & 33.7 & 34.2 & 19.4 & 19.9\\
concat22 & 23.8 & 24.4 & \textbf{34.5}& 35.0 & \textbf{20.2} & 20.8 \\
\midrule
\multicolumn{3}{l}{\bf independent encoders}\\
s-hier & 23.5 & 24.0 & 33.5 & 34.0 & 18.9 & 19.5\\
s-hier-to-2  & 23.8 & 24.3 & 34.2 & 34.8 & 19.2 & 19.7\\
s-t-hier & 23.1 & 23.6 & 33.1 & 33.6 & 19.3 & 20.0\\
\midrule
\multicolumn{3}{l}{\bf with weight tying}\\
s-hier.tied & 23.6 & 24.1 & 33.7 & 34.2 & 19.7 & 20.3\\
s-hier-to-2.tied   & \textbf{24.2} & 24.8 & 34.1 & 34.7 & 20.1 & 20.7 \\
s-t-hier.tied & 23.5 & 24.0 & 33.9 & 34.5 & 19.4 & 20.0\\
\midrule 
\multicolumn{3}{l}{\bf Transformer-based models}\\
baseline & - & 24.6 & - & 35.4 & - & 21.1\\
concat21 & - & 24.8 & - & 35.3 & - & \textbf{21.8}\\
concat22 & - & 24.4 & - & 36.0 & - & 21.3\\
\cite{Voita2018} & - & \textbf{25.3} & - & \textbf{36.5} & - & 21.7 \\
\bottomrule
\end{tabular}
\caption{English$\to$German BLEU scores on newstest2017, newstest2018 and all sentence pairs from our pronoun test set. Case-sensitive and case-insensitive (uncased) scores are reported. Higher is better, and the best scores are marked in bold.}\label{Tab:bleu}
\end{table*}

\section{Evaluation} \label{sec:evaluation}

The BLEU scores in Table \ref{Tab:bleu} show a moderate improvement for most context-aware systems. This suggests that the architectural changes for the context-aware models do not degrade overall translation quality. The contrastive evaluation on our test set on the other hand shows a clear increase in the accuracy of pronoun translation: The best model {\em s-hier-to-2.tied} achieves a total of +16 percentage points accuracy on the test set over the baseline, see Table \ref{Tab:pairs}.

\begin{table}
\centering
\begin{tabular}{lcccc}
\toprule
& & \multicolumn{3}{c}{reference pronoun}\\
           & total & {\em es} & {\em er} & {\em sie} \\ \cmidrule{2-5}
baseline   &  0.44 & 0.85 & 0.17 & 0.31\\
concat22   &  0.53 & 0.84 & 0.32 & 0.42\\
\midrule
\multicolumn{3}{l}{\bf independent encoders}\\
s-hier & 0.43 & 0.80 & 0.20 & 0.29 \\
s-hier-to-2 & 0.55 & 0.84 & 0.41 & 0.40\\
s-t-hier    & 0.52 & 0.88 & 0.32 & 0.36\\
\midrule
\multicolumn{3}{l}{\bf with weight tying}\\
s-hier.tied & 0.47 & 0.85 & 0.30 & 0.26 \\
s-hier-to-2.tied & \textbf{0.60} & 0.87 & \textbf{0.45} &\textbf{0.48}\\
s-t-hier.tied    & 0.56 & 0.86 & 0.39 & 0.42 \\
\midrule
\multicolumn{3}{l}{\bf Transformer-based models}\\
baseline & 0.47 & 0.81 & 0.22 & 0.38 \\
concat21 & 0.48 & 0.88 & 0.26 & 0.31 \\
concat22 & 0.49 & \textbf{0.91} & 0.20 & 0.36 \\
\cite{Voita2018} & 0.49 & 0.84 & 0.23 & 0.39 \\
\bottomrule
\end{tabular}\textbf{}
\caption{Accuracy on contrastive test set (N=4000 per pronoun) with regard to reference pronoun.}\label{Tab:pairs}
\end{table}

\begin{table}
\begin{center}
\begin{tabular}{lcc}
\toprule
& \multicolumn{2}{c}{antecedent location}\\
 & intrasegmental  & external \\ \cmidrule{2-3}
baseline & 0.57& 0.41 \\
concat22 & 0.58 & 0.51 \\
\midrule
\multicolumn{3}{l}{\bf independent encoders}\\
s-hier & 0.58 & 0.39 \\
s-hier-to-2 & 0.63 & 0.53 \\
s-t-hier & 0.52 & 0.52\\
\midrule
\multicolumn{3}{l}{\bf with weight tying}\\
s-hier.tied &  0.56 & 0.45 \\
s-hier-to-2.tied  & 0.65  & \textbf{0.58} \\
s-t-hier.tied  & 0.57 & 0.55 \\
\midrule
\multicolumn{3}{l}{\bf Transformer-based models}\\
baseline & 0.70 & 0.41 \\
concat21 & 0.67 & 0.44 \\
concat22 & 0.56 & 0.47 \\
\cite{Voita2018} & \textbf{0.75} & 0.43 \\
\bottomrule
\end{tabular}
\caption{Accuracy on contrastive test set with regard to antecedent location (within segment vs.\ outside segment).}\label{Tab:segmental}
\end{center}
\end{table}

Table \ref{Tab:segmental} shows that context-aware models perform better than the baseline when the antecedent is outside the current sentence. 
In our experiments, all context-aware models consider one preceding sentence as context.
The evaluation according to the distance of the antecedent in Table \ref{Tab:antedist} confirms that the subset of sentences with antecedent distance 1 benefits most from the tested context-aware models (up to +20 percentage points accuracy). 
However, we note two surprising patterns:

\begin{itemize}
\item For inter-sentential anaphora, the performance of all systems, including the baseline, improves with increasing antecedent distance.
\item Context-aware systems that consider one preceding sentence also improve on intra-sentential anaphora, and on pronouns whose antecedent is outside the context window.
\end{itemize}

\begin{table*}
\begin{center}
\begin{tabular}{lccccc}
\toprule
 & \multicolumn{5}{c}{antecedent distance} \\
                  & 0 	 & 1 	& 2    & 3    & $>$3 \\ \cmidrule{2-6}
baseline &          0.57 & 0.38 & 0.47 & 0.52 & 0.67 \\
concat22 &  		0.58 & 0.50 & 0.51 & 0.51 & 0.69 \\
\midrule
\multicolumn{3}{l}{\bf independent encoders}\\
s-hier & 			0.58 & 0.36 & 0.42  & 0.46 & 0.61 \\
s-hier-to-2 &		0.63 & 0.51 & 0.54 & 0.60 & 0.70 \\
s-t-hier & 			0.52 & 0.49 & \textbf{0.57} & \textbf{0.61} & 0.71 \\
\midrule
\multicolumn{3}{l}{\bf with weight tying}\\
s-hier.tied & 			0.56	& 0.43 & 0.46 & 0.49 & 0.67 \\
s-hier-to-2.tied  & 0.65 & \textbf{0.58} & 0.55 & 0.55 & \textbf{0.75} \\
s-t-hier.tied & 	0.57 & 0.54 & 0.56 & 0.59 & 0.72 \\
\midrule
\multicolumn{3}{l}{\bf Transformer-based models}\\
baseline & 0.70 & 0.38 & 0.45 & 0.49 & 0.65 \\
concat21 & 0.67 & 0.42 & 0.45 & 0.47 & 0.66 \\
concat22 & 0.56 & 0.44 & 0.53 & 0.54 & 0.74 \\
\cite{Voita2018} & \textbf{0.75} & 0.39 & 0.48 & 0.54 & 0.66 \\
\bottomrule
\end{tabular}
\caption{Accuracy on contrastive test set with regard to antecedent distance of antecedent (in sentences).}\label{Tab:antedist}
\end{center}
\end{table*}

The first observation can be explained by the distribution of German pronouns in the test set. The further away the antecedent, the higher the percentage of {\em it}$\rightarrow${\em es} cases, which are the majority class, and thus the class that will be predicted most often if evidence for other classes is lacking. We speculate that this is due to our more permissive extraction heuristics for {\em it}$\rightarrow${\em es}.

\begin{table*}
\begin{center}
\begin{tabular}{ll}
\toprule
source sentence with antecedent & {\em What's with the door?}\\
target sentence with antecedent & {\em Was ist mit der T\"ur?}\\
source context & {\em \textbf{It} won't open.}\\
reference context & {\em \textbf{Sie} geht nicht auf.}\\
source sentence & {\em - Is \textbf{it} locked?} \\
reference sentence & {\em - Ist \textbf{sie} abgeschlossen?} \\ \midrule
contrastive 1 & {\em - Ist \textbf{er} abgeschlossen?} \\
contrastive 2 & {\em - Ist \textbf{es} abgeschlossen?}\\
\bottomrule
\end{tabular}
\caption{Example where 1) antecedent distance is  $>$1 and 2) the context given contains another pronoun as an additional hint.}\label{Ex:farAnte}
\end{center}
\end{table*}

We attribute the second observation to the existence of coreference chains where the preceding sentence contains a pronoun that refers to the same nominal antecedent as the pronoun in the current sentence. Consider the example in Table \ref{Ex:farAnte}: The nominal antecedent of {\em it} in the current sentence is {\em door}, {\em T\"ur} in German with feminine gender. The nominal antecedent occurs two sentences before the current sentence, but the German sentence in between contains the pronoun {\em sie}, which is a useful signal for the context-aware models, even though they cannot know the nominal antecedent.

Note that only models aware of target-side context can benefit from such circumstances: The {\em s-hier} models as well as the Transformer model by \cite{Voita2018} only see source side context, which results in lower accuracy if the distance to the antecedent is $>$1, see Table \ref{Tab:antedist}.

While such coreference chains complicate the interpretation of the results, we note that improvements on inter-sentential anaphora with antecedent distance $>1$ are relatively small (compared to distance 1), and that performance is still relatively poor (especially for the minority classes {\em er} and {\em sie}). We encourage evaluation of wider-context models on this subset, which is still large thanks to the size of the full test set.

Regarding the comparison of different context-aware architectures, our results demonstrate the effectiveness of parameter sharing between the main encoder (or decoder) and the contextual encoder.
We observe an improvement of 5 percentage points from {\em s-hier-to-2} to {\em s-hier-to-2.tied}, and 4 percentage points from {\em s-t-hier} to {\em s-t-hier.tied}.
Context encoders introduce a large number of extra parameters, while inter-sentential context is only relevant for a relatively small number of predictions.
We hypothesize that the training signal is thus too weak to train a strong contextual encoder in an end-to-end fashion without parameter sharing.
Our results also confirm the finding by \citet{Bawden2017} that multi-encoder architectures, specifically {\em s-hier-to-2(.tied)}, can outperform a simple concatenation system in the translation of coreferential pronouns.

The Transformer-based models perform strongest on pronouns with intra-segmental antecedent, outperforming the recurrent baseline by 9--18 percentage points.
This is likely an effect of increased model depth and the self-attentional architecture in this set of experiments.
The model by \cite{Voita2018} only uses source context, and outperforms the most comparable RNN system, {\em s-hier.tied}.
However, the Transformer-based {\em concat22} slightly underperforms the RNN-based {\em concat22}, and we consider it future research how to better exploit target context with Transformer-based models.

\section{Conclusions}
We present a large-scale test suite to specifically test the capacity of NMT models to translate pronouns correctly. The test set contains 12,000 difficult cases of pronoun translations from English {\em it} to its German counterparts {\em er, sie} and {\em es}, extracted automatically from OpenSubtitles \citep{Lison2016}.

We evaluate recently proposed context-aware models on our test set. Even though the increase in BLEU score is moderate for all context-aware models, the improvement in the translation of pronouns is considerable: The best model ({\em s-hier-to-2.tied}) achieves a +16 percentage points gain in accuracy over the baseline.

Our experiments confirm the importance of careful architecture design, with multi-encoder architectures outperforming a model
that simply concatenates context sentences. We also demonstrate the effectiveness of parameter sharing between encoders of a context-aware model.

We hope the test set will prove useful for empirically validating novel architectures for context-aware NMT.
So far, we have only evaluated models that consider one sentence of context, but the nominal antecedent is more distant for a sizable proportion of the test set, and the evaluation of variable-size context models \cite{Wang2017,DBLP:conf/acl/HaffariM18} is interesting future work.

\section*{Acknowledgements}
We  are  grateful  to  the  Swiss  National  Science Foundation (SNF) for supporting the project \mbox{CoNTra} (grant number 105212\_169888).

\bibliography{wmt18}
\bibliographystyle{acl_natbib}

\end{document}